\documentclass{ecai} 
\usepackage{latexsym}
\usepackage{amssymb}
\usepackage{amsmath}
\usepackage{amsthm}
\usepackage{booktabs}
\usepackage{enumitem}
\usepackage{graphicx}
\usepackage{color}

\usepackage{multirow}
\usepackage[linesnumbered,ruled,vlined]{algorithm2e}
\newcommand{\BibTeX}{B\kern-.05em{\sc i\kern-.025em b}\kern-.08em\TeX}

\begin{document}

%%%%%%%%%%%%%%%%%%%%%%%%%%%%%%%%%%%%%%%%%%%%%%%%%%%%%%%%%%%%%%%%%%%%%%%%

\begin{frontmatter}

%%% Use this command to specify your submission number.
%%% In doubleblind mode, it will be printed on the first page.

\paperid{0380} 

%%% Use this command to specify the title of your paper.

\title{Enhancing Local Search for MaxSAT with Deep Differentiation Clause Weighting}

%%% Use this combinations of commands to specify all authors of your 
%%% paper. Use \fnms{} and \snm{} to indicate everyone's first names 
%%% and surname. This will help the publisher with indexing the 
%%% proceedings. Please use a reasonable approximation in case your 
%%% name does not neatly split into "first names" and "surname".
%%% Specifying your ORCID digital identifier is optional. 
%%% Use the \thanks{} command to indicate one or more corresponding 
%%% authors and their email address(es). If so desired, you can specify
%%% author contributions using the \footnote{} command.

\author[A]{\fnms{Menghua}~\snm{Jiang}}
\author[A]{\fnms{Haokai}~\snm{Gao}}
\author[A]{\fnms{Shuhao}~\snm{Chen}}
\author[A,B]{\fnms{Yin}~\snm{Chen}\thanks{Corresponding Author. Email: ychen@scnu.edu.cn.}} 

\address[A]{School of Computer Science, South China Normal University, Guangzhou, China}
\address[B]{School of Artificial Intelligence, South China Normal University, Foshan, China}

%%% Use this environment to include an abstract of your paper.

\begin{abstract}
Partial Maximum Satisfiability (PMS) and Weighted Partial Maximum Satisfiability (WPMS) generalize Maximum Satisfiability (MaxSAT), with broad real-world applications. Recent advances in Stochastic Local Search (SLS) algorithms for solving (W)PMS have mainly focused on designing clause weighting schemes. However, existing methods often fail to adequately distinguish between PMS and WPMS, typically employing uniform update strategies for clause weights and overlooking critical structural differences between the two problem types. In this work, we present a novel clause weighting scheme that, for the first time, updates the clause weights of PMS and WPMS instances according to distinct conditions. This scheme also introduces a new initialization method, which better accommodates the unique characteristics of both instance types. Furthermore, we propose a decimation method that prioritizes satisfying unit and hard clauses, effectively complementing our proposed clause weighting scheme. Building on these methods, we develop a new SLS solver for (W)PMS named DeepDist. Experimental results on benchmarks from the anytime tracks of recent MaxSAT Evaluations show that DeepDist outperforms state-of-the-art SLS solvers. Notably, a hybrid solver combining DeepDist with TT-Open-WBO-Inc surpasses the performance of the MaxSAT Evaluation 2024 winners, SPB-MaxSAT-c-Band and SPB-MaxSAT-c-FPS, highlighting the effectiveness of our approach. The code is available at \url{https://github.com/jmhmaxsat/DeepDist}
\end{abstract}

\end{frontmatter}

%%%%%%%%%%%%%%%%%%%%%%%%%%%%%%%%%%%%%%%%%%%%%%%%%%%%%%%%%%%%%%%%%%%%%%%%

\section{Introduction}
The Maximum Satisfiability (MaxSAT) problem is an optimization extension of the Boolean Satisfiability (SAT) problem and is widely applied in fields such as computer science, mathematical logic, and artificial intelligence. The core objective of MaxSAT is to find a variable assignment for a given Conjunctive Normal Form (CNF) formula that maximizes the number of satisfied clauses. The Partial Maximum Satisfiability (PMS) problem extends this by dividing clauses into hard and soft clauses, requiring all hard clauses to be satisfied while maximizing the number of satisfied soft clauses. Furthermore, the Weighted Partial Maximum Satisfiability (WPMS) problem extends PMS by assigning different weights to soft clauses, aiming to maximize the total weight of satisfied soft clauses while still satisfying all hard clauses. Many real-world optimization tasks involve both hard constraints and soft preferences with varying priorities, making (W)PMS a more appropriate model than SAT or MaxSAT for such problems. Typical applications include timetabling \cite{lemos2020minimal}, planning \cite{bonet2019learning}, routing \cite{khadilkar2022solving}, and combinatorial testing \cite{ansotegui2022incomplete}.

Currently, mainstream algorithms for (W)PMS can be categorized into two types based on whether they guarantee finding the optimal solution: exact algorithms \cite{borchers1998two,zhang2003exact} and anytime algorithms \cite{cai2014tailoring,cai2020old}. These are the two primary tracks evaluated in the recent MaxSAT Evaluations (MSEs)\footnote{\url{https://maxsat-evaluations.github.io}}. Exact algorithms are capable of finding the optimal solution by the end of the solving process, providing a proof of its optimality, or proving that no solution exists. Common exact algorithms include Branch-and-Bound (BnB) algorithms \cite{cherif2020understanding, li2025improving}, as well as methods based on multiple calls to SAT solvers, generally referred to as SAT-based algorithms \cite{fu2006solving,ansotegui2009solving, ansotegui2010new,ansotegui2013sat,narodytska2014maximum,martins2014open,ansotegui2017wpm3,berg2020abstract}. SAT-based algorithms were first proposed in \cite{fu2006solving} for PMS and later extended to WPMS in \cite{ansotegui2009solving, ansotegui2010new}. These algorithms have garnered significant attention in recent years and have undergone continuous refinement in various studies \cite{berg2019core, nadel2019anytime,niskanen2021enabling,ihalainen2024unifying}.

Anytime algorithms for solving (W)PMS are primarily stochastic local search (SLS) algorithms \cite{cha1997local,lei2018solving,cai2020old,zheng2022bandmaxsat,zheng2023farsighted, chu2023nuwls,chu2024enhancing,zheng2024rethinking,zheng2025integrating,histls}. These algorithms are particularly effective in many large-scale industrial problems where time costs are high. Their goal is not to find the optimal solution, but to obtain a high-quality solution within a reasonable time. In recent years, with the introduction and development of various techniques, anytime algorithms have seen significant improvements and now rival exact algorithms in industrial applications \cite{cai2020old}. Currently, most leading anytime solvers employ hybrid approaches, where SLS algorithm is first used for preprocessing, followed by the integration of SAT-based methods \cite{lei2021satlike, chu2022nuwls, chu2023nuwls-c,jiang2024nuwls,jin2024combining}. This combination has dominated all four anytime tracks of recent MSEs \cite{chu2024enhancing}. In this paper, we focus specifically on the SLS components of solvers for (W)PMS.

SLS algorithms for (W)PMS iteratively select a variable and flip its value (from \textit{True} to \textit{False}, or vice versa), following a greedy strategy to satisfy all hard clauses while maximizing the number or total weight of satisfied soft clauses. The main challenge is that it can easily fall into local optima, where flipping any variable does not improve the number or total weight of satisfied soft clauses. To address this issue, recent research has focused on designing advanced clause weighting schemes \cite{cai2014tailoring,cai2020old,chu2023nuwls,jiang2024raslite,chu2024enhancing,zheng2024rethinking}. The main idea behind these schemes is to increase the weight of falsified clauses at local optima, encouraging the algorithm to prioritize satisfying these clauses in subsequent local searches.

However, despite the significant progress achieved by existing clause weighting schemes, they generally fail to make a deeper distinction between PMS and WPMS instances. These methods often use the same conditions to update the weights of soft clauses, failing to leverage the structural differences between these types of instances. This unified weighting strategy may overlook the fundamental difference between PMS and WPMS: in PMS, soft clauses are treated equally, whereas in WPMS, the priority of soft clauses must be considered. On the other hand, assigning higher initial weights to hard clauses than to soft clauses at the early stage of the search is reasonable, as it helps the solver quickly find a feasible solution. However, continuing to prioritize hard clause weights even after a feasible solution has been found may be suboptimal, as this initialization can cause the solver to repeatedly follow the same search trajectory in each local search iteration, thereby reducing search efficiency.

In this work, we propose a carefully designed clause weighting scheme called Deep-Weighting, which is the first to update the weights of soft clauses of PMS and WPMS instances based on distinct conditions. This scheme explicitly accounts for the differing weight characteristics of these two types of instances, effectively guiding the SLS solver to search in directions better aligned with their structural properties. Furthermore, Deep-Weighting introduces a novel initialization method that diversifies initial search paths and mitigates the risk of search stagnation. Notably, when solving PMS instances, Deep-Weighting resets the initial weights of hard clauses to zero once the first feasible solution has been found. To the best of our knowledge, this is the first approach to explicitly propose initializing hard clause weights to zero.

We further propose a new decimation method called UnH-Decimation, which is used in conjunction with Deep-Weighting to initialize PMS instances. This method aims to satisfy more hard clauses when generating the initial solution. Specifically, UnH-Decimation prioritizes satisfying unit clauses and hard clauses. When contradictory unit clauses exist, it assigns values to the associated variables based on the best solution found in the most recent local search. It is worth noting that this idea was first introduced in DeciLS \cite{cai2017decimation}, but was gradually abandoned in subsequent methods. In this work, we revisit and apply this strategy within the framework we propose.

Building upon the aforementioned methods, we propose a new SLS solver, DeepDist. Experimental results on benchmarks from the anytime tracks of recent MSEs show that DeepDist outperforms state-of-the-art SLS solvers. Additionally, we combine DeepDist with the typical SAT-based solver TT-Open-WBO-Inc \cite{nadel2019anytime} to form a hybrid solver, denoted as DeepDist-c, as many anytime solvers do \cite{cai2020old, zheng2022bandmaxsat, chu2023nuwls, chu2024enhancing, zheng2024rethinking}. The experiments demonstrate that DeepDist-c outperforms the state-of-the-art hybrid solvers, SPB-MaxSAT-c-Band and SPB-MaxSAT-c-FPS, which were the winners of the MaxSAT Evaluation 2024.

The main contributions of this work are as follows:
\begin{itemize}
\item We propose a novel clause weighting scheme, Deep-Weighting, which updates clause weights for PMS and WPMS instances based on distinct conditions. This scheme also introduces a new initialization method that better accommodates the unique characteristics of both types of instances.
\item We propose a new decimation method, UnH-Decimation, which prioritizes satisfying more hard clauses when generating initial solutions. UnH-Decimation is used in conjunction with Deep-Weighting to initialize PMS instances.
\item Building upon these methods, we develop a new SLS solver for (W)PMS, named DeepDist. Extensive experiments show that DeepDist outperforms state-of-the-art SLS solvers.
\end{itemize}

\section{Preliminaries}
\label{2-Preliminaries}
Given a set of Boolean variables \( V = \{ x_1, x_2, \dots, x_n \} \), a literal is defined as either a variable \( x_i \) (positive literal) or its negation \( \neg x_i \) (negative literal). A positive literal \( x_i \) is considered satisfied if \( x_i \) is assigned the value \textit{True}, while a negative literal \( \neg x_i \) is satisfied when \( x_i \) is \textit{False}. A clause is a disjunction of literals, i.e., \( c_i = l_{i1} \lor l_{i2} \lor \dots \lor l_{ik} \), where \( k \) is the number of literals in the clause \( c_i \). A clause is satisfied if at least one of its literals is satisfied; otherwise, it is falsified. A CNF formula \( F \) is the conjunction of all its clauses, i.e., \( F = c_1 \land c_2 \land \dots \land c_m \), where \( m \) is the total number of clauses in the formula. The set of variables used in the formula \( F \) is denoted as \( V(F) \). An assignment \( \alpha: V(F) \rightarrow \{0, 1\} \) maps each variable to a Boolean value, with 1 representing \textit{True} and 0 representing \textit{False}. A complete assignment refers to an assignment where values have been assigned to all variables in \( V(F) \). In this context, all assignments discussed are assumed to be complete.

Given a CNF formula \( F \), the objective of MaxSAT is to find an assignment that satisfies the maximum number of clauses in \( F \). A variant of MaxSAT, called PMS, classifies clauses into hard and soft, aiming to satisfy all hard clauses while maximizing the number of satisfied soft clauses. WPMS, an extension of PMS, assigns a weight to each soft clause and seeks an assignment that satisfies all hard clauses while maximizing the total weight of satisfied soft clauses. Obviously, PMS can be viewed as a special case of WPMS in which every soft clause has a weight of one.

Clause weighting schemes are frequently utilized in SLS solvers for MaxSAT, where each clause is assigned two types of weights: the original weight \( w_{\textit{org}}(c) \), which is the weight assigned to \( c \) in the CNF formula and used to compute the \textit{cost}, and the dynamic weight \( w(c) \), hereafter simply referred to as weight, which is updated by the clause weighting scheme during the search process and used to compute the variable \textit{score}. Given a CNF formula \( F \) and an assignment \( \alpha \), we say that \( \alpha \) is a solution to \( F \). The solution \( \alpha \) is considered feasible if it satisfies all the hard clauses in \( F \). The \textit{cost} of a feasible solution \( \alpha \), denoted as \( \textit{cost}(\alpha) \), is defined as the number of falsified soft clauses in the PMS, and the total original weight of falsified soft clauses in the WPMS under \( \alpha \). If \( \alpha \) is infeasible, its \textit{cost} is assigned a value of \( +\infty \) for convenience. Furthermore, a solution \( \alpha_1 \) is said to be better than another solution \( \alpha_2 \) if \( \textit{cost}(\alpha_1) < \textit{cost}(\alpha_2) \).

In SLS solvers for (W)PMS, flipping a variable involves changing its Boolean value from \textit{True} to \textit{False}, or vice versa. For a variable \( x \), the \textit{gain} of \( x \), denoted as \( \textit{gain}(x) \), represents the total weight of the falsified clauses that would become satisfied if \( x \) is flipped. The \textit{loss} of \( x \), denoted as \( \textit{loss}(x) \), is the total weight of the satisfied clauses that would become falsified if \( x \) is flipped. The \textit{score} of \( x \), denoted as \( score(x) \), indicates the net change in the total weight of satisfied clauses resulting from flipping \( x \), and is computed as \( score(x) = \textit{gain}(x) - \textit{loss}(x) \). If \( score(x) > 0 \), then \( x \) is considered a \textit{GoodVar}, as flipping it is expected to result in a positive net change in the total weight of satisfied clauses, making it a beneficial move in the search process.

To improve performance, SLS solvers often incorporate a restart strategy when the current solution does not show any improvement after a predetermined number of steps. This strategy involves reinitializing the current assignment and resetting the clause weights. At the start of each local search iteration, SLS solvers must initialize \( w(c) \) for every clause \( c \), where \( w_{init}(c) \) represents the initial weight of the clause. Furthermore, \( avg_{soft} \) denotes the average original weight of the soft clauses. Proper initialization of \( w_{init}(c) \) is vital for effectively steering the search and enhancing solver performance.

% The notation \( w_{init}(c) \) represents the initial weight of \( c \) at the start of each local search iteration, and \( avg_{soft} \) denotes the average original weight of the soft clauses.

\section{Related Works}
\label{03-RelatedWorks}
Clause weighting is a classical technique used in local search algorithms for SAT and MaxSAT \cite{frank1996weighting, cha1997local}. In this approach, each clause is assigned a weight that is dynamically adjusted during the search, which helps the solver escape local optimum. Early studies typically assigned a very large initial weight to hard clauses and treated them as soft clauses, either statically \cite{cha1997local} or dynamically \cite{thornton1998dynamic, thornton2002two}.

A representative recent work is Dist \cite{cai2014tailoring}, which was the first to introduce the concepts of hard $score$ and soft $score$ to separately evaluate the impact of flipping a variable on the satisfaction of hard and soft clauses. Dist has inspired a series of variants, such as DistUP \cite{cai2016new}, CCEHC \cite{luo2017ccehc}, and NuDist \cite{lei2020nudist}, all of which have led to significant improvements over earlier SLS solvers.

SATLike \cite{lei2018solving} introduced a novel clause weighting scheme called Weighting-PMS, which assigns differentiated weights to hard and soft clauses while imposing a uniform upper bound on the weight of each soft clause. By integrating the feasible solution priority strategy with Weighting-PMS, SATLike3.0 \cite{cai2020old} was developed. This clause weighting scheme has also been adopted by BandMaxSAT \cite{zheng2022bandmaxsat}, MaxFPS \cite{zheng2023farsighted}, and BandHS \cite{zheng2025integrating}.

Recently, NuWLS \cite{chu2023nuwls} introduced an improved clause weighting scheme by assigning appropriate initial weights to soft clauses and adopting distinct conditions for updating the weights of both hard and soft clauses. This resulting scheme was named Dist-Weighting. Building upon this, USW-LS \cite{chu2024enhancing} proposed a unified soft clause weighting scheme called Unified-SW, which increases the weights of all soft clauses in a feasible local optimum—regardless of whether they are satisfied—while preserving their hierarchical structure. Additionally, SPB-MaxSAT \cite{zheng2024rethinking} transferred the concept of Soft Conflict Pseudo-Boolean (SPB) constraints from exact algorithms to the clause weighting system of local search, giving rise to SPB-Weighting, which offers new perspectives for clause weighting in MaxSAT local search solvers.

However, despite the aforementioned progress, designing effective clause weighting strategies for SLS solvers remains a significant challenge. This is because the initial or adjusted weights may not accurately reflect the importance or satisfiability difficulty of hard or soft clauses, potentially leading to disruptions in the search space.

\section{Methodology}
\label{04-Methodology}
In this section, we first present the proposed Deep Differentiation Clause Weighting (Deep-Weighting) scheme, which also introduces a new initialization method tailored to the unique characteristics of both instance types. Next, we introduce the Unit and Hard Clauses Decimation (UnH-Decimation) method, which prioritizes the satisfaction of unit and hard clauses. Together with Deep-Weighting, UnH-Decimation effectively initializes PMS instances. Finally, we outline the main procedure of DeepDist.

\subsection{Deep Differentiation Clause Weighting Scheme}
In this subsection, we introduce Deep-Weighting, which operates as follows:

\textbf{Initialization of Clause Weights:} At the start of each round of local search, Deep-Weighting initializes the weight of each clause as follows:
\begin{itemize}
      \item For each clause \( c \) in \textbf{PMS} instances: If the solution found in the previous round of local search is infeasible, set \( w_{init}(c) := 1 \) for hard clauses and \( w_{init}(c) := 0 \) for soft clauses. Otherwise, set \( w_{init}(c) := 0 \) for hard clauses and \( w_{init}(c) := 1 \) for soft clauses.
      \item For each clause \( c \) in \textbf{WPMS} instances: If the solution found in the previous round of local search is infeasible, set \( w_{init}(c) := 1 \) for hard clauses and \( w_{init}(c) := 0 \) for soft clauses. Otherwise, set \( w_{init}(c) := 1 \) for hard clauses and \( w_{init}(c) := \frac{w_{org}(c)}{avg_{soft}} \) for soft clauses.
\end{itemize}

\textbf{Update of Clause Weights:} When the search encounters a local optimum \( \alpha \), the clause weights are updated as follows:

\begin{itemize}
      \item For each hard clause \( c \) in \textbf{(W)PMS} instances: If \( \alpha \) is infeasible, set \( w(c) := w(c) + h_{inc} \).
      \item For each soft clause \( c \) in \textbf{PMS} instances: If the first feasible solution has been found and the \textit{cost} of the current solution \( \alpha \) isn't smaller than that of the best solution found so far \( \alpha^* \) (i.e., \( cost(\alpha) \geq cost(\alpha^*) \)), set \( w(c) := \delta \times (w(c) + \frac{w_{org}(c)}{avg_{soft}}) \).
      \item For each soft clause \( c \) in \textbf{WPMS} instances: If \( \alpha \) is feasible, set \( w(c) := \delta \times (w(c) + \frac{w_{org}(c)}{avg_{soft}}) \).
\end{itemize}

Our clause weighting scheme is the first to update the weights of soft clauses in PMS and WPMS instances based on distinct conditions, while also introducing a novel initialization method that sets the initial weight of hard clauses to zero after the SLS solver finds the first feasible solution for the PMS instance. To the best of our knowledge, this is the first initialization strategy with such a design. Below, we provide some discussion on these ideas.

At the start of the first round of local search, regardless of whether it is a PMS or WPMS instance, the $scores$ of any variable do not consider soft clauses, as their weights are set to zero. At this stage, the solver focuses on finding the first feasible solution that satisfies all hard clauses without considering soft clauses, essentially solving a SAT problem. Deep-Weighting helps facilitate the rapid discovery of a feasible solution by increasing the weight of unsatisfied hard clauses, thus laying the foundation for the subsequent search.

The search continues until flipping any variable no longer improves the current solution, indicating that it has reached a local optimum. At this point, Deep-Weighting updates the soft clause weights of PMS and WPMS instances under distinct conditions, unlike previous methods that apply identical conditions to both. For PMS instances, if the first feasible solution has been found and the $cost$ of the current solution \( \alpha \) is not smaller than the $cost$ of the best solution found so far \( \alpha^* \) (i.e., \( cost(\alpha) \geq cost(\alpha^*) \)), Deep-Weighting updates the weights of all soft clauses. For WPMS instances, if the current solution is feasible, Deep-Weighting updates the weights of all soft clauses. Notably, for the amount of weight update for the clauses, we adopt SPB-Weighting \cite{zheng2024rethinking}, which increases the dynamic weights proportionally according to the current weights.

\textbf{Why we set different conditions: }Before the first feasible solution is found, the solver should focus only on the hard clauses. At this stage, assigning weights only to the hard clauses is reasonable, as it helps quickly locate a feasible solution. However, once a feasible solution is found, continuing to use this strategy for PMS becomes inappropriate. Although soft clauses in PMS are equally important, they differ in their difficulty of being satisfied. If we continue to update soft clause weights only when the current solution is feasible, the solver will tend to favor satisfying the easier clauses (since state-of-the-art clause weighting schemes use unified soft clause weighting). In contrast, when the $cost$ of the current solution is smaller than that of the best solution found so far, we do not update the soft clause weights. This encourages the solver to continue searching in the neighborhood of the current solution, helping to avoid redundant search paths. For WPMS, it is reasonable to shift the focus to soft clauses after a feasible solution has been found. In this case, the solver tends to prioritize satisfying soft clauses with higher weights because we also adopt the unified soft clause weighting strategy, which preserves the original weight hierarchy and aligns with our solving objective.

When the current solution fails to improve after a certain number of steps (a parameter set by the solver), the solver adopts a restart strategy. During the local search restart, in addition to regenerating the initial solution, the weights assigned to each clause are also reset. At this point, Deep-Weighting also initializes the soft clause weights for PMS and WPMS instances under different conditions. For WPMS instances, if the solver found a feasible solution in the previous round of local search, Deep-Weighting sets \( w_{init}(c) = 1 \) for hard clauses and \( w_{init}(c) = \frac{w_{org}(c)}{avg_{soft}} \) for soft clauses. For PMS instances, if the solver found a feasible solution in the previous round of local search, Deep-Weighting sets \( w_{init}(c) = 0 \) for hard clauses and \( w_{init}(c) = 1 \) for soft clauses. This is the second key difference from other methods. Deep-Weighting initially sets the weight of soft clauses higher than that of hard clauses, which may temporarily move the solver away from a better solution. However, the solver will later take advantage of the higher hard clause weights after the update, using new search paths to progress toward a better solution, ultimately expanding the solver’s search space.

\subsection{Unit and Hard Clauses Decimation Method}
Since Deep-Weighting assigns higher initial weights to soft clauses than to hard clauses for PMS instances, we propose the UnH-Decimation method to generate the initial solution, aiming to prevent the solver from straying too far from better solutions. UnH-Decimation prioritizes satisfaction of unit and hard clauses. It improves upon UP-Decimation \cite{cai2020old}, typically producing an initial solution that satisfies more hard clauses, thus helping mitigate solver divergence from promising regions of the search space.

Unit Propagation (UP) has been widely used in local search algorithms for SAT and MaxSAT due to its ability to maintain instance satisfiability \cite{heras2011impact, abrame2012inference}. Recently, UP has also been employed to generate initial solutions for SLS algorithms solving (W)PMS, in a process known as UP-Decimation \cite{cai2020old}. The proposed UnH-Decimation method is inspired by UP-Decimation, but differs in two key ways. Experiments presented in the next section demonstrate that UnH-Decimation is more effective when integrated into DeepDist.

The procedure of UnH-Decimation is shown in Algorithm \ref{algorithm1}, and we explain the two main differences between UnH-Decimation and UP-Decimation as follows. Firstly, when there are two contradictory hard unit clauses (denoted by the variable related to the two contradictory unit clauses as \( v \)), UnH-Decimation assigns \( v \) the value of the best solution found by the most recent local search (denoted as \( prev\_ls\_best \)), i.e., \( v := prev\_ls\_best[v] \). In contrast, UP-Decimation randomly assigns a value to \( v \) and continues the unit propagation process. The strategy of assigning values based on the best solution found by the latest local search was first introduced in DeciLS \cite{cai2017decimation}, and UnH-Decimation reintroduces this idea, as the most recent local search typically satisfies more hard clauses. Secondly, when there are no unit clauses, UnH-Decimation selects unassigned variables from the falsified hard clauses for propagation, whereas UP-Decimation randomly selects unassigned variables. This strategy allows UnH-Decimation to satisfy more hard clauses than UP-Decimation.

% It is important to note that UnH-Decimation is specifically designed to complement the initialization method of Deep-Weighting, reducing some of the randomness in the search process. Therefore, it may not be a universally applicable approach, as this reduction in randomness could limit the diversity of initial solutions, potentially having a detrimental effect on the solver's overall performance.
It is important to note that UnH-Decimation is specifically designed to complement the initialization method of Deep-Weighting, reducing some of the randomness in the search process. Thus, it may not be universally applicable, as less randomness can limit solution diversity and potentially harm overall performance.

\begin{algorithm}[t]
\caption{UnH-Decimation}
\label{algorithm1}
\KwIn{A PMS instance $F$}
\KwOut{An assignment of variables in $F$}

\While{$\exists$ unassigned variables}
{    
    \If{$\exists$ \text{unit clauses}}
    { 
        \If{$\exists$ \text{contradictory unit clauses}}
        {
            $v :=$ the variable related to the two contradictory unit clauses\;
            
            assign $v$ to $prev\_ls\_best[v]$ and simplify $F$\;
        }
        \Else
        {
           $c :=$ \textnormal{a randomly picked unit clause}\;
           \textnormal{satisfy $c$ and simplify $F$}\;
        }
    } 
    % \ElseIf{$\exists$ \text{ soft unit clauses}}
    % {
    %     \If{$\exists$ \text{contradictory unit clauses}}
    %     {
    %         $v :=$ the variable related to the two contradictory unit clauses\;

    %         $value :=$ \textnormal{the value of $v$ in $\alpha$}\;
            
    %         assign $x$ to $prev\_ls\_best[x]$, simplify $F$ accordingly\;
    %     }
    %     \Else
    %     {
    %        $c :=$ \textnormal{a randomly picked soft unit clause}\;
    %        \textnormal{satisfy $c$ and simplify $F$}\;
    %     }
    % }       
    \ElseIf{$\exists$ \text{ falsified hard clauses}}
    {
        $c :=$ \textnormal{a randomly picked falsified hard clause}\;
        
        \textnormal{satisfy $c$ and simplify $F$}\;
    }
    \Else
    {
        $v :=$ \textnormal{a randomly picked unassigned variable}\;
        
        \textnormal{assign $v$ randomly and simplify $F$}\;
    }
}

\Return{\textnormal{the resulting assignment to \( Var(F) \)}}\;
\end{algorithm}

\subsection{The DeepDist Algorithm}

We present the pseudo-code of DeepDist in Algorithm~\ref{algorithm2} and describe its main procedure as follows. Let $\alpha^*$ denote the best solution found so far, and $\textit{cost}^*$ its corresponding $cost$. During the local search process, $\alpha$ represents the current assignment. Initially, $\alpha^*$ is set to an empty assignment, and $\textit{cost}^*$ is initialized to $+\infty$.

DeepDist iteratively invokes the local search process until the \textit{cutoff time} is reached (lines 2--22). In each local search iteration, DeepDist first generates an initial complete assignment $\alpha$ using UnH-Decimation (line 4) for PMS instances, and UP-Decimation \cite{cai2020old} (line 6) for WPMS instances. It then initializes clause weights (line 7) and inserts all variables $x$ with $\textit{score}(x) > 0$ into the set $GoodVars$. After initialization, DeepDist enters the main local search phase (lines 9-22). During the search, whenever a solution with a lower $cost$ than the current best $\textit{cost}^*$ is found, the best assignment $\alpha^*$ and $\textit{cost}^*$ are updated accordingly.

In each local search step, DeepDist selects a variable to flip to improve the current solution. When the search falls into local optimum, DeepDist adopts the Best from Multiple Selections (BMS) strategy \cite{cai2015balance} to determine which variable to flip. This strategy is essential because traversing all variables in the set $GoodVars$ can be computationally expensive. BMS mitigates this by randomly selecting \( k \) variables from $GoodVars$ (where \( k \) is predefined) and choosing the one with the highest $score$ (lines 13–14). If $GoodVars$ is empty—signaling that the search encounters a local optimum, DeepDist updates the clause weights according to Deep-Weighting (line 16). It then randomly selects a falsified clause \( c \), and from it chooses the variable with the highest $score$ for flipping (lines 17–21). 

Finally, if a feasible solution is found after the \textit{cutoff time} has been reached, DeepDist reports \(\alpha^*\) and \(\textit{cost}^*\); otherwise, it reports ``No feasible solution found''. 
\begin{algorithm}[!t]
\caption{DeepDist}
\label{algorithm2}
\KwIn{A (W)PMS instance $F$, \textit{cutoff time}}
\KwOut{The best solution found and its $cost$, or ``No feasible solution found''.}
$\alpha^* := \emptyset$; $cost^* := +\infty$\;
\While{running time $<$ cutoff time}
{
    \If{$F$ is a PMS instance}
    {
        $\alpha :=$ an initial assignment by UnH-Decimation\;  
    }
    \Else
    {
        $\alpha :=$ an initial assignment by UP-Decimation\;  
    }
    Initialize clause weights by Deep-Weighting\;
    $L := 10000000$\;
    \For{step = 0; step $<$ L; step++}
    {
        \If{$\alpha$ is feasible \textbf{and} $cost(\alpha) < cost^*$}
        {
            $\alpha^* := \alpha$; $cost^* := cost(\alpha)$\; 
            $L := step + 10000000$;
        }
        \If{$GoodVars : \{x \mid \text{score}(x) > 0\} \neq \emptyset$}
        {
            $v :=$ a variable in $GoodVars$ selected by the BMS strategy\;
        }
        \Else
        {
            update clause weights by Deep-Weighting\;
            \If{$\exists$ falsified hard clauses}
            {
                $c :=$ a randomly picked falsified hard clause\;
            }
            \Else
            {
                $c :=$ a randomly picked falsified soft clause\;
            }
            
            $v :=$ the variable with highest $score$ in $c$\;
        }
        
        $\alpha := \alpha$ with $v$ flipped\;
    }
}
\lIf{$\alpha^* \neq \emptyset$}{\textbf{return} $\alpha^*$  and $cost^*$}
\lElse{\textbf{return} ``No feasible solution found''}
\end{algorithm}

\section{Experiments}
\label{05-Experiments}
In this section, we first introduce the benchmarks, competitors, parameter settings, and experimental setup used in our experiments. We then present the results on both unweighted and weighted benchmarks from the anytime tracks of the MSEs held between 2020 and 2024 to evaluate the effectiveness of our DeepDist algorithm. Finally, we conduct ablation studies to assess the contributions of individual components within DeepDist.

\subsection{Experimental Preliminaries}

\subsubsection{Benchmarks}
Our experiments are conducted on 10 benchmarks, including both unweighted and weighted benchmarks from the anytime tracks of the MSEs held from 2020 to 2024. Note that we denote the benchmark containing all the PMS / WPMS instances from the anytime tracks of MSE 2020 as \texttt{PMS\_2020} / \texttt{WPMS\_2020}, and similarly for subsequent years.

\subsubsection{State-of-the-art Competitors}
In the first experiment, we compare DeepDist against 8 state-of-the-art SLS solvers, whose source codes are publicly available\footnote{\url{http://lcs.ios.ac.cn/~caisw/Code/maxsat}},%
\footnote{\url{https://github.com/JHL-HUST}},%
\footnote{\url{https://github.com/filyouzicha}}. The parameter settings used in our experiments follow those specified in the respective original papers. These solvers have been used as the SLS components in hybrid systems and have achieved notable results in the anytime tracks of recent MSEs, specifically:

{\renewcommand{\labelitemi}{--} % 使用横线代替黑点
\begin{itemize}
    \item \textbf{SATLike3.0}~\cite{cai2020old}: runner-up of the weighted anytime tracks of MSE 2020 and MSE 2021.
    \item \textbf{BandMaxSAT}~\cite{zheng2022bandmaxsat}, \textbf{MaxFPS}~\cite{zheng2023farsighted}: third place of the weighted anytime track of MSE 2022.
    \item \textbf{NuWLS}~\cite{chu2023nuwls}: champion of all anytime tracks of MSE 2022.
    \item \textbf{USW-LS}~\cite{chu2024enhancing}: champion of all anytime tracks of MSE 2023.
    \item \textbf{SPB-MaxSAT}~\cite{zheng2024rethinking}: champion of all anytime tracks of MSE 2024.
    \item \textbf{BandHS}~\cite{zheng2025integrating}, \textbf{NuWLS-BandHS}~\cite{zheng2025integrating}: recently proposed SLS solvers.
\end{itemize}
}

In the second experiment, we combine DeepDist with TT-Open-WBO-Inc \cite{nadel2019anytime}, which is based on Open-WBO-Inc \cite{joshi2019open}. We compare the resulting hybrid solver DeepDist-c with the state-of-the-art hybrid solvers SPB-MaxSAT-c-Band and SPB-MaxSAT-c-FPS, which respectively won the unweighted and weighted anytime tracks of MSE 2024. We use their source code from MSE 2024 \footnote{\url{https://maxsat-evaluations.github.io/2024/descriptions.html}}. Due to their outstanding results in the recent MSEs, we only select them as the baseline for hybrid solvers, while disregarding other effective solvers such as Loandra \cite{berg2019core}, TT-Open-WBO-Inc \cite{nadel2019anytime} and NuWLS-c-2023 \cite{chu2023nuwls-c}.

\subsubsection{Parameter Settings.}
DeepDist is controlled by three parameters: (1) \( h_{inc} \), which denotes the increment of hard clause weights in the clause weighting scheme; (2) \( \delta \), the multiplicative factor used to increase \( h_{inc} \); and (3) \( k \), the number of samples used in the BMS strategy. The parameter settings are as follows: for PMS instances, \( h_{inc} = 1 \), \( \delta = 1.00072 \), and \( k = 53 \); for WPMS instances, \( h_{inc} = 28 \), \( \delta = 1.001 \), and \( k = 97 \). These values are adopted directly from the baseline solver SPB-MaxSAT, and no further parameter tuning was conducted.

\subsubsection{Experimental Setup.}
DeepDist is implemented in C++ and compiled using the \texttt{g++} compiler with the \texttt{-O3} optimization flag. To ensure fairness and consistency, all experiments involving DeepDist and its competitors were conducted on a unified server under the same experimental environment. The server is equipped with an Intel\textsuperscript{\textregistered} Xeon\textsuperscript{\textregistered} Gold 6230 CPU @ 2.10GHz (80 cores) and 377\,GB of RAM, running the Ubuntu 22.04 Linux operating system.

We followed the evaluation methodology used in the anytime tracks of MSEs. Each solver was executed once per instance, and for each run, we recorded the best solution found along with the time required to find it, measured using the \texttt{runsolver} tool~\cite{roussel2011controlling} (version 3.4.0). The cutoff time was set to 60 and 300 CPU seconds, consistent with the rules of the anytime tracks of recent MSEs.

To assess solver performance, we use two commonly adopted metrics: ``$\#win.$'' and ``$\#score$''. The ``$\#win.$'' metric counts the number of instances for which a solver achieves the best solution among all competitors listed in the same table. This metric is widely used in the comparison of SLS solvers~\cite{cai2020old,zheng2022bandmaxsat,zheng2023farsighted,chu2023nuwls,chu2024enhancing,zheng2024rethinking,zheng2025integrating}.

The ``$\#score$'' metric, adopted in the anytime tracks of MSEs, evaluates each solver on a per-instance basis. For an instance \(i\), the $score$ of a solver is defined as 0 if it fails to produce a feasible solution. Otherwise, the $score$ is computed as $\textit{score}(\textit{solver}, i) = \frac{cost_{best} + 1}{\textit{cost}(\alpha^*) + 1}$, 
where \(\alpha^*\) is the feasible solution found by the solver, \(\textit{cost}(\alpha^*)\) denotes its $cost$, and \(\textit{cost}_{best}\) is the lowest $cost$ obtained for that instance among all solvers. The overall performance of a solver on a benchmark is measured by the average of its per-instance $score$, represented by ``$\#score$''. In all result tables, the best values are highlighted in bold.

\begin{table}[t]
\footnotesize
\centering
\caption{Comparison of DeepDist and USW-LS.}
\label{table-USW-LS}
\resizebox{\linewidth}{!}{%
\begin{tabular}{lrrrrrrr}
\bottomrule
\multirow{2}{*}{Benchmark} & \multirow{2}{*}{\#inst.} & & 
\multicolumn{2}{c}{DeepDist} & & \multicolumn{2}{c}{USW-LS} \\ 
\cline{4-5} \cline{7-8} 
 & & & $\#win.$ & $\#score$ & & $\#win.$ & $\#score$ \\ 
\hline
\multicolumn{8}{l}{Time Limit: 60s} \\
\texttt{PMS\_2020} & 262 & & \textbf{167} & \textbf{0.7891} & & 151 & 0.7842 \\
\texttt{PMS\_2021} & 155 & & \textbf{101} & \textbf{0.7799} & & 96  & 0.7786 \\
\texttt{PMS\_2022} & 179 & & \textbf{115} & \textbf{0.8022} & & 98  & 0.7808 \\
\texttt{PMS\_2023} & 179 & & \textbf{110} & \textbf{0.8105} & & 95  & 0.7811 \\
\texttt{PMS\_2024} & 216 & & \textbf{145} & \textbf{0.8234} & & 132 & 0.8111 \\
\texttt{WPMS\_2020} & 253 & & \textbf{144} & \textbf{0.8341} & & 125 & 0.8077 \\
\texttt{WPMS\_2021} & 151 & & \textbf{83}  & \textbf{0.7802} & & 62  & 0.7525 \\
\texttt{WPMS\_2022} & 197 & & \textbf{106} & \textbf{0.7801} & & 88  & 0.7580 \\
\texttt{WPMS\_2023} & 160 & & \textbf{78}  & \textbf{0.7730} & & 77  & 0.7190 \\
\texttt{WPMS\_2024} & 229 & & \textbf{134} & 0.8148 & & 121 & \textbf{0.8149} \\
\hline
\multicolumn{8}{l}{Time Limit: 300s} \\
\texttt{PMS\_2020} & 262 & & \textbf{169} & \textbf{0.7930} & & 159 & 0.7846 \\
\texttt{PMS\_2021} & 155 & & \textbf{100} & 0.7780 & & 96  & \textbf{0.7793} \\
\texttt{PMS\_2022} & 179 & & \textbf{124} & \textbf{0.8032} & & 105 & 0.7830 \\
\texttt{PMS\_2023} & 179 & & \textbf{124} & \textbf{0.8101} & & 94  & 0.7961 \\
\texttt{PMS\_2024} & 216 & & \textbf{143} & \textbf{0.8185} & & 140 & 0.8157 \\
\texttt{WPMS\_2020} & 253 & & \textbf{157} & \textbf{0.8352} & & 124 & 0.8175 \\
\texttt{WPMS\_2021} & 151 & & \textbf{92}  & \textbf{0.7863} & & 62  & 0.7517 \\
\texttt{WPMS\_2022} & 197 & & \textbf{112} & \textbf{0.7790} & & 93  & 0.7619 \\
\texttt{WPMS\_2023} & 160 & & \textbf{79}  & \textbf{0.7702} & & 71  & 0.7256 \\
\texttt{WPMS\_2024} & 229 & & \textbf{145} & \textbf{0.8172} & & 113 & 0.8121 \\
\toprule
\end{tabular}
}
\end{table}

\subsection{Comparison with SLS Solvers}
To assess the effectiveness of DeepDist, we present a detailed comparison with USW-LS and SPB-MaxSAT, both of which adopt a unified soft clause weighting strategy that significantly enhances their performance compared to other solvers. As shown in Tables~\ref{table-USW-LS} and~\ref{table-SPB-MaxSAT}, DeepDist consistently outperforms these solvers across almost all benchmarks under both the 60s and 300s time limits. For PMS benchmarks, the number of ``$\#win.$'' instances of DeepDist is 2.14\%--31.91\% (resp. 1.80\%--9.85\%) more than USW-LS (resp. SPB-MaxSAT), while for WPMS benchmarks the improvement is 1.30\%--48.39\% (resp. 10.53\%--62.71\%). In terms of the ``$\#score$'' metric, DeepDist achieves gains of 0.17\%--3.76\% (resp.  0.24\%--3.61\%) over USW-LS (resp. SPB-MaxSAT) for PMS, and 0.63\%--7.51\% (resp. 1.46\%--5.36\%) for WPMS. Although it slightly underperforms on a few datasets---for instance, the ``$\#score$'' on \texttt{PMS\_2021} is 0.0013 lower than USW-LS, and on \texttt{PMS\_2023} is 0.0017 lower than SPB-MaxSAT---these differences are marginal. Overall, DeepDist demonstrates robust and superior performance across diverse benchmark scenarios.

% Under the 60-second time limit, DeepDist surpasses USW-LS and SPB-MaxSAT in the ``$\#win.$'' metric by 5.21\%--17.35\% and 1.92\%--9.85\%, respectively, and in the ``$\#score$'' metric by 0.17\%--3.76\% and 0.95\%--3.61\%. With a 300-second time limit, DeepDist achieves improvements of 2.14\%--31.91\% and 1.80\%--6.62\% in ``$\#win.$'', and 0.34\%--2.58\% and 0.24\%--2.70\% in ``$\#score$'', respectively. Although it slightly underperforms on a few datasets---for instance, the ``$\#score$'' on \texttt{PMS\_2021} is 0.0013 lower than USW-LS, and on \texttt{PMS\_2023} is 0.0017 lower than SPB-MaxSAT---these differences are marginal. Overall, DeepDist demonstrates robust and superior performance across diverse benchmark scenarios.

\begin{figure}[t]
\centering
\includegraphics[width=\linewidth]{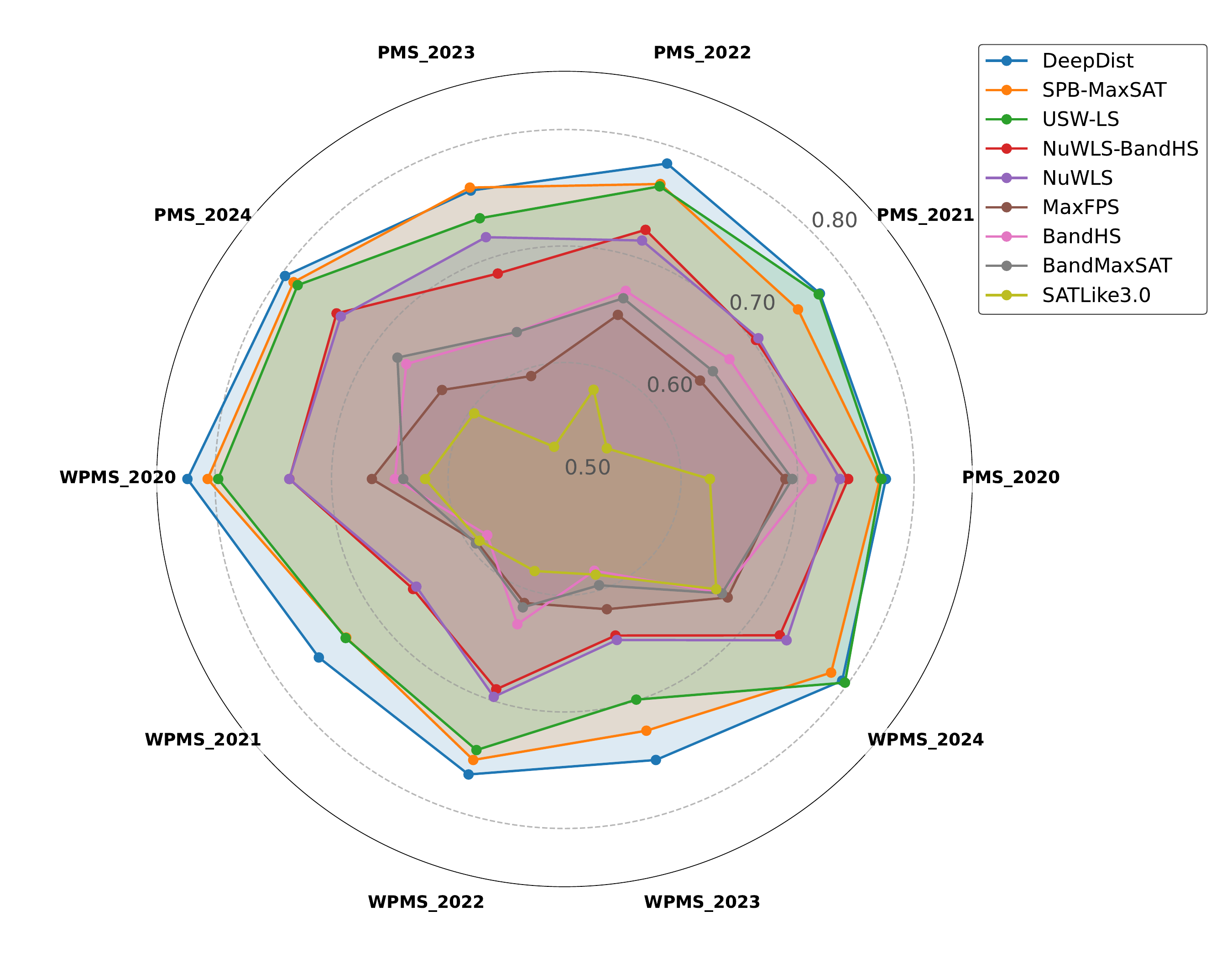} 
\caption{Comparison of DeepDist and baseline SLS solvers within 60s based on the ``\textit{\#score}'' metric}
\label{fig:60s}
\end{figure}

\begin{figure}[t]
\centering
\includegraphics[width=\linewidth]{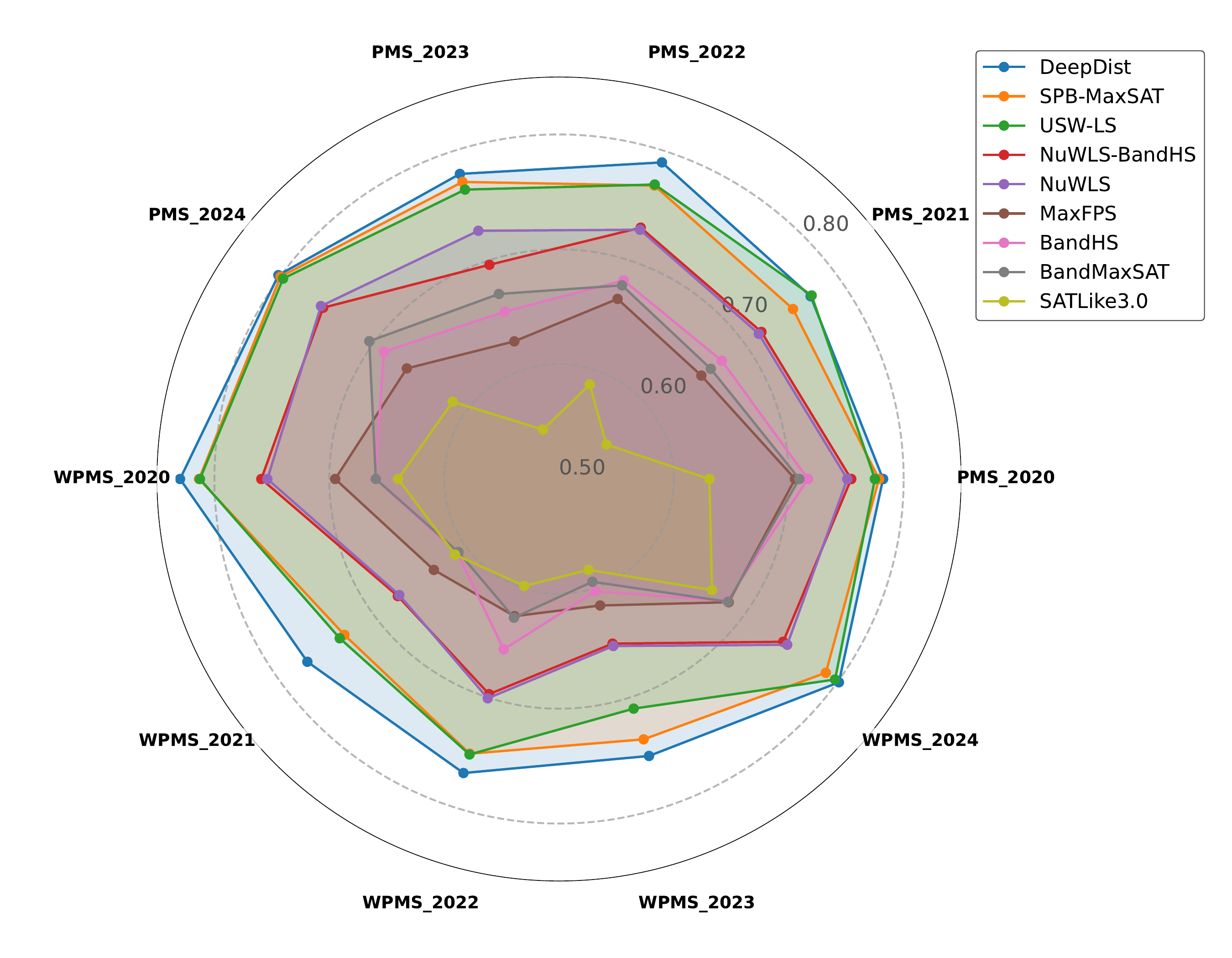} 
\caption{Comparison of DeepDist and baseline SLS solvers within 300s based on the ``\textit{\#score}'' metric}
\label{fig:300s}
\end{figure}

The improvements achieved by DeepDist become even more evident under the 300s time limit, primarily because the core idea of Deep-Weighting is to minimize the solver’s tendency to revisit previously explored search paths, thereby expanding the effective search space. As the allowed runtime increases, this advantage becomes increasingly pronounced. Furthermore, we provide a comparison between DeepDist and all baseline SLS solvers based on the ``$\#score$'' metric. As illustrated in Figures~\ref{fig:60s} and~\ref{fig:300s}, DeepDist consistently outperforms its competitors. Notably, the proposed method achieves more pronounced leading performance in most WPMS benchmarks. In summary, the proposed method integrates Deep-Weighting with UnH-Decimation, helping DeepDist to outperform competitors.

\begin{table}[t]
\footnotesize
\centering
\caption{Comparison of DeepDist and SPB-MaxSAT.}
\label{table-SPB-MaxSAT}
\resizebox{\linewidth}{!}{%
\begin{tabular}{lrrrrrrr}
\bottomrule
\multirow{2}{*}{Benchmark} & \multirow{2}{*}{\#inst.} & & 
\multicolumn{2}{c}{DeepDist} & & \multicolumn{2}{c}{SPB-MaxSAT} \\ 
\cline{4-5} \cline{7-8} 
 & & & $\#win.$ & $\#score$ & & $\#win.$ & $\#score$ \\ 
\hline
\multicolumn{8}{l}{Time Limit: 60s} \\
\texttt{PMS\_2020} & 262 & & \textbf{167} & \textbf{0.7882} & & 159 & 0.7808 \\
\texttt{PMS\_2021} & 155 & & \textbf{100} & \textbf{0.7781} & & 95  & 0.7510 \\
\texttt{PMS\_2022} & 179 & & \textbf{110} & \textbf{0.7969} & & 106 & 0.7751 \\
\texttt{PMS\_2023} & 179 & & \textbf{106} & 0.7888 & & 104 & \textbf{0.7905} \\
\texttt{PMS\_2024} & 216 & & \textbf{145} & \textbf{0.8188} & & 132 & 0.8077 \\
\texttt{WPMS\_2020} & 253 & & \textbf{145} & \textbf{0.8392} & & 122 & 0.8216 \\
\texttt{WPMS\_2021} & 151 & & \textbf{77}  & \textbf{0.7818} & & 67  & 0.7528 \\
\texttt{WPMS\_2022} & 197 & & \textbf{109} & \textbf{0.7791} & & 82  & 0.7661 \\
\texttt{WPMS\_2023} & 160 & & \textbf{82}  & \textbf{0.7735} & & 72  & 0.7473 \\
\texttt{WPMS\_2024} & 229 & & 123 & \textbf{0.8113} & & \textbf{133} & 0.7996 \\
\hline
\multicolumn{8}{l}{Time Limit: 300s} \\
\texttt{PMS\_2020} & 262 & & \textbf{170} & \textbf{0.7912} & & 167 & 0.7867 \\
\texttt{PMS\_2021} & 155 & & 94  & \textbf{0.7758} & & \textbf{101} & 0.7571 \\
\texttt{PMS\_2022} & 179 & & \textbf{119} & \textbf{0.8021} & & 114 & 0.7810 \\
\texttt{PMS\_2023} & 179 & & \textbf{110} & \textbf{0.8008} & & 108 & 0.7954 \\
\texttt{PMS\_2024} & 216 & & \textbf{145} & \textbf{0.8198} & & 136 & 0.8178 \\
\texttt{WPMS\_2020} & 253 & & \textbf{148} & \textbf{0.8387} & & 134 & 0.8222 \\
\texttt{WPMS\_2021} & 151 & & \textbf{96}  & \textbf{0.7921} & & 59  & 0.7518 \\
\texttt{WPMS\_2022} & 197 & & \textbf{115} & \textbf{0.7849} & & 91  & 0.7659 \\
\texttt{WPMS\_2023} & 160 & & \textbf{84}  & \textbf{0.7715} & & 76  & 0.7564 \\
\texttt{WPMS\_2024} & 229 & & \textbf{141} & \textbf{0.8173} & & 124 & 0.8030 \\
\toprule
\end{tabular}
}
\end{table}

\begin{table}[t]
\footnotesize
\centering
\caption{Comparison of DeepDist-c with SPB-MaxSAT-c-Band (SPB-MaxSAT-c-B) and SPB-MaxSAT-c-FPS (SPB-MaxSAT-c-F).}
\label{table-DeepDist-c}
\resizebox{\linewidth}{!}{%
\begin{tabular}{lrrrrrrrrrr}
\bottomrule
\multirow{2}{*}{Benchmark} & \multirow{2}{*}{\#inst.} & & 
\multicolumn{2}{c}{DeepDist-c} & & \multicolumn{2}{c}{SPB-MaxSAT-c-B} & & \multicolumn{2}{c}{SPB-MaxSAT-c-F} \\ 
\cline{4-5} \cline{7-8} \cline{10-11}
 & & & 60s & 300s & & 60s & 300s & & 60s & 300s \\ 
\hline
\texttt{PMS\_2020}  & 253 
& & \textbf{0.9169}    & \textbf{0.9168}  
& & \textbf{0.9169}    & 0.9138  
& & 0.9090    & 0.9134\\

\texttt{PMS\_2021}  & 151 
& & \textbf{0.9080}    & \textbf{0.9066}   
& & 0.8973    & 0.8873  
& & 0.8757    & 0.8879\\

\texttt{PMS\_2022}  & 197 
& & 0.8782    & \textbf{0.8842}   
& & \textbf{0.8868}    & 0.8828  
& & 0.8564    & 0.8574\\

\texttt{PMS\_2023}  & 160 
& & 0.8548    & \textbf{0.8514}   
& & \textbf{0.8638}    & 0.8506  
& & 0.8199    & 0.8492\\

\texttt{PMS\_2024}  & 229 
& & \textbf{0.8990}    & \textbf{0.9071}   
& & 0.8927    & 0.8908  
& & 0.8666    & 0.8658\\ 

\texttt{WPMS\_2020}  & 253 
& & \textbf{0.8929}    & 0.8965   
& & 0.8898    & \textbf{0.9055}  
& & 0.8748    & 0.8874\\

\texttt{WPMS\_2021}  & 151 
& & 0.8595    & \textbf{0.8778}   
& & 0.8600    & 0.8717  
& & \textbf{0.8745}    & 0.8714\\

\texttt{WPMS\_2022}  & 197 
& & 0.8193    & 0.8311   
& & \textbf{0.8389}    & \textbf{0.8445}  
& & 0.8296    & 0.8349\\

\texttt{WPMS\_2023}  & 160 
& & \textbf{0.9036}    & 0.9116   
& & 0.8918    & \textbf{0.9140}  
& & 0.8787    & 0.8874\\

\texttt{WPMS\_2024}  & 229 
& & \textbf{0.9137}    & \textbf{0.9191}  
& & 0.8969    & 0.9151  
& & 0.8907    & 0.9136\\ 
\toprule
\end{tabular}
}
\end{table}

\subsection{Comparison with Hybrid Solvers}
The comparison results between DeepDist-c, SPB-MaxSAT-c-Band, and SPB-MaxSAT-c-FPS are presented in Table~\ref{table-DeepDist-c}. All three solvers combine their SLS components (i.e., DeepDist, SPB-MaxSAT \cite{zheng2024rethinking}, BandMaxSAT \cite{zheng2022bandmaxsat}, and FPS \cite{zheng2023farsighted}) with the SAT-based solver TT-Open-WBO-Inc \cite{nadel2019anytime}. Since the MSEs only compare solvers in the anytime tracks using the ``\textit{\#score}'' metric, we follow the same practice in this paper.

For PMS instances, DeepDist-c outperforms SPB-MaxSAT-c-Band, the champion of the unweighted anytime track of MSE 2024—particularly under the 300s time limit, where it achieves the best performance across all benchmarks. This strong performance is primarily attributed to the innovative hard clause initial weight zeroing strategy introduced in Deep-Weighting, which effectively expands the solver’s search space. For WPMS instances, DeepDist-c also demonstrates better performance, outperforming both SPB-MaxSAT-c-Band and SPB-MaxSAT-c-FPS overall, winning half of the evaluated metrics. Interestingly, SPB-MaxSAT-c-FPS, the champion of the weighted anytime track of MSE 2024, ranks last in this comparison. This may be because DeepDist-c performs even better on instances where SPB-MaxSAT-c-FPS typically excels, thereby reducing the latter’s relative advantage.

\begin{table}[t]
\footnotesize
\centering
\caption{Comparison of DeepDist and DeepDist-alt1.}
\label{table-DeepDist-alt1}
\resizebox{\linewidth}{!}{%
\begin{tabular}{lrrrrrrr}
\bottomrule
\multirow{2}{*}{Benchmark} & \multirow{2}{*}{\#inst.} & & 
\multicolumn{2}{c}{DeepDist} & & \multicolumn{2}{c}{DeepDist-alt1} \\ 
\cline{4-5} \cline{7-8} 
 & & & $\#win.$ & $\#score$ & & $\#win.$ & $\#score$ \\ 
\hline
\multicolumn{8}{l}{Time Limit: 60s} \\
\texttt{PMS\_2020} & 262 & & \textbf{170} & \textbf{0.7925} & & 159 & 0.7893 \\
\texttt{PMS\_2021} & 155 & & \textbf{103} & \textbf{0.7810} & & 97 & 0.7677 \\
\texttt{PMS\_2022} & 179 & & \textbf{117} & \textbf{0.8017} & & 110 & 0.7966 \\
\texttt{PMS\_2023} & 179 & & 109 & \textbf{0.8113} & & \textbf{110} & 0.8095 \\
\texttt{PMS\_2024} & 216 & & \textbf{147} & \textbf{0.8218} & & 139 & 0.8209 \\
\texttt{WPMS\_2020} & 253 & & \textbf{150} & \textbf{0.8390} & & 129 & 0.8256 \\
\texttt{WPMS\_2021} & 151 & & \textbf{81} & 0.7824 & & 73 & \textbf{0.7864} \\
\texttt{WPMS\_2022} & 197 & & \textbf{109} & 0.7806 & & 93 & \textbf{0.7853} \\
\texttt{WPMS\_2023} & 160 & & \textbf{95} & \textbf{0.7763} & & 79 & 0.7491 \\
\texttt{WPMS\_2024} & 229 & & \textbf{130} & 0.8132 & & 129 & \textbf{0.8185} \\
\hline
\multicolumn{8}{l}{Time Limit: 300s} \\
\texttt{PMS\_2020} & 262 & & \textbf{180} & \textbf{0.7941} & & 166 & 0.7932 \\
\texttt{PMS\_2021} & 155 & & \textbf{105} & \textbf{0.7793} & & 98 & 0.7708 \\
\texttt{PMS\_2022} & 179 & & \textbf{121} & \textbf{0.8029} & & 113 & 0.7952 \\
\texttt{PMS\_2023} & 179 & & \textbf{122} & \textbf{0.8121} & & 100 & 0.8092 \\
\texttt{PMS\_2024} & 216 & & \textbf{149} & 0.8224 & & 138 & \textbf{0.8277} \\
\texttt{WPMS\_2020} & 253 & & \textbf{160} & \textbf{0.8417} & & 124 & 0.8224 \\
\texttt{WPMS\_2021} & 151 & & \textbf{90} & \textbf{0.7914} & & 67 & 0.7856 \\
\texttt{WPMS\_2022} & 197 & & \textbf{113} & 0.7844 & & 92 & \textbf{0.7861} \\
\texttt{WPMS\_2023} & 160 & & \textbf{97} & \textbf{0.7772} & & 75 & 0.7446 \\
\texttt{WPMS\_2024} & 229 & & \textbf{139} & 0.8141 & & 127 & \textbf{0.8213} \\
\toprule
\end{tabular}
}
\end{table}

\subsection{Ablation Study}
To evaluate the effectiveness and rationality of the proposed Deep-Weighting and UnH-Decimation methods, we compare the DeepDist algorithm with three of its variants: DeepDist-alt1, DeepDist-alt2, and DeepDist-alt3. Specifically, DeepDist-alt1 modifies the condition used in Deep-Weighting by swapping the criteria for updating the soft clause weights in PMS and WPMS instances, in order to verify the adaptability and soundness of the proposed weighting scheme for different instance types. DeepDist-alt2, when solving PMS instances, does not reset the initial weights of hard clauses to 0 after obtaining the first feasible solution; instead, it follows the same initialization strategy as the baseline solver SPB-MaxSAT, setting all soft clause weights to 1. DeepDist-alt3 replaces the proposed UnH-Decimation method with UP-Decimation to generate the initial solution for PMS instances.

\begin{table}[t]
\footnotesize
\centering
\caption{Comparison of DeepDist, DeepDist-alt2 and DeepDist-alt3.}
\label{table-DeepDist-alt2-3}
\resizebox{\linewidth}{!}{%
\begin{tabular}{lrrrrrrrrrr}
\bottomrule
\multirow{2}{*}{Benchmark} & \multirow{2}{*}{\#inst.} & & 
\multicolumn{2}{c}{DeepDist} & & \multicolumn{2}{c}{DeepDist-alt2} & & \multicolumn{2}{c}{DeepDist-alt3} \\ 
\cline{4-5} \cline{7-8} \cline{10-11} 
 & & & $\#win.$ & $\#score$ & & $\#win.$ & $\#score$ & & $\#win.$ & $\#score$ \\ 
\hline
\multicolumn{11}{l}{Time Limit: 60s} \\
\texttt{PMS\_2020} & 262 & & \textbf{151} & \textbf{0.7862} & & 149 & 0.7797 & & 141 & 0.7794 \\
\texttt{PMS\_2021} & 155 & & \textbf{99} & \textbf{0.7761} & & 89 & 0.7482 & & 81 & 0.7610 \\
\texttt{PMS\_2022} & 179 & & \textbf{111} & \textbf{0.7966} & & 98 & 0.7664 & & 86 & 0.7775 \\
\texttt{PMS\_2023} & 179 & & 99 & \textbf{0.7897} & & \textbf{101} & 0.7876 & & 82 & 0.7738 \\
\texttt{PMS\_2024} & 216 & & \textbf{134} & \textbf{0.8127} & & 133 & 0.8108 & & 113 & 0.8021 \\
\hline
\multicolumn{11}{l}{Time Limit: 300s} \\
\texttt{PMS\_2020} & 262 & & \textbf{161} & \textbf{0.7906} & & 159 & 0.7855 & & 147 & 0.7824 \\
\texttt{PMS\_2021} & 155 & & 89 & \textbf{0.7714} & & 89 & 0.7507 & & \textbf{92} & 0.7681 \\
\texttt{PMS\_2022} & 179 & & \textbf{118} & \textbf{0.7969} & & 110 & 0.7708 & & 93 & 0.7817 \\
\texttt{PMS\_2023} & 179 & & \textbf{98} & 0.7941 & & 93 & \textbf{0.8047} & & 91 & 0.7854 \\
\texttt{PMS\_2024} & 216 & & 128 & 0.8129 & & \textbf{135} & 0.8138 & & 123 & \textbf{0.8150} \\
\toprule
\end{tabular}
}
\end{table}

The comparison results between DeepDist and its three variants are summarized in Tables~\ref{table-DeepDist-alt1} and~\ref{table-DeepDist-alt2-3}. The results show that DeepDist outperforms DeepDist-alt1 on the majority of benchmarks, validating the effectiveness of using different conditions to update soft clause weights in PMS and WPMS instances and confirming the good compatibility of this strategy with the corresponding instance types. Compared with DeepDist-alt2 and DeepDist-alt3, DeepDist also demonstrates superior performance on almost all benchmarks, indicating the effectiveness of initializing the weights of hard clauses to zero. Moreover, the proposed UnH-Decimation method, which prioritizes both unit clauses and hard clauses during initial solution generation, provides more reasonable initial assignments for PMS instances than strategies that consider only unit clauses.

\section{Conclusion}
\label{06-Conclusion}
Existing clause weighting schemes have achieved notable success, but they often fail to make a deeper distinction between PMS and WPMS, thereby overlooking the structural characteristics unique to each instance type. In this work, we propose a novel clause weighting scheme that, for the first time, updates clause weights separately for these two categories of instances based on distinct conditions. Additionally, the scheme introduces an innovative initialization strategy that resets the weights of hard clauses to zero after the first feasible solution is found. We further propose a decimation method that prioritizes satisfying unit clauses and hard clauses, which is designed to work in harmony with the proposed initialization strategy. Based on these innovations, we present a new SLS solver named DeepDist. Experimental results show that DeepDist outperforms existing SLS solvers on standard (W)PMS benchmarks and further enhances the performance of hybrid solvers designed for (W)PMS.

Future work will explore the relationship between hard clause and soft clause increments to achieve a more balanced weighting scheme, as the increment for hard clauses currently remains a fixed constant.

%%% Use this command to include your bibliography file.
\bibliography{mybibfile}

\end{document}